\documentclass[letterpaper, 10 pt, conference]{ieeeconf}  

\IEEEoverridecommandlockouts                              

\title{\LARGE \bf Integrating Human-Provided Information Into Belief State Representation Using Dynamic Factorization}
\author{
  \textbf{Rohan Chitnis \hspace{16mm} Leslie Pack Kaelbling \hspace{8mm} Tom\'as Lozano-P\'erez}\\\\
  MIT Computer Science and Artificial Intelligence Laboratory\\
  \texttt{\{ronuchit, lpk, tlp\}@mit.edu}
  \thanks{\hspace*{-1em}Presented at the 2018 IEEE/RSJ International Conference on Intelligent Robots and Systems (IROS), Madrid, Spain.}
}

\usepackage{multicol}
\usepackage{color}
\usepackage{amssymb}
\usepackage{amsmath}
\usepackage{tikz}
\usetikzlibrary{arrows,decorations.markings}

\usepackage{booktabs}
\usepackage{graphicx}

\usepackage[font=small]{caption}
\usepackage{subcaption}

\usepackage{algorithm2e}
\usepackage{indentfirst}
\usepackage{float}
\usepackage{mathtools}

\newenvironment{tightlist}%
{\begin{list}{$\bullet$}{%
    \setlength{\topsep}{0in}
    \setlength{\partopsep}{0in}
    \setlength{\itemsep}{0in}
    \setlength{\parsep}{0in}
    \setlength{\leftmargin}{1.5em}
    \setlength{\rightmargin}{0in}
    \setlength{\itemindent}{-.1in}
}
}%
{\end{list}
}

\newcommand{\figref}[1]{Figure~\ref{#1}}
\newcommand{\algref}[1]{Algorithm~\ref{#1}}
\newcommand{\tabref}[1]{Table~\ref{#1}}

\newtheorem{defn}{Definition}

\DeclarePairedDelimiterX{\infdivx}[2]{(}{)}{%
  #1\;\delimsize\|\;#2%
}
\newcommand{\kl}{D_{KL}\infdivx}
\newcommand{\js}{D_{JS}\infdivx}

\newcommand{\Obj}{\mathcal{O}}
\newcommand{\F}{\mathcal{F}}
\newcommand{\I}{\mathcal{I}}
\newcommand{\G}{\mathcal{G}}
\renewcommand{\P}{\mathcal{P}}
\newcommand{\St}{\mathcal{S}}
\newcommand{\A}{\mathcal{A}}
\newcommand{\T}{\mathcal{T}}

\newcommand{\V}{\mathcal{V}}

\newcommand{\U}{\mathcal{U}}
\newcommand{\pomdp}{{\sc pomdp}}
\newcommand{\Obs}{\Omega}

\begin{document}

\bibliographystyle{IEEEtran}

\maketitle
\thispagestyle{empty}
\pagestyle{empty}
\setlength{\textfloatsep}{8pt}

\begin{abstract}
  In partially observed environments, it can be useful for a human to
  provide the robot with declarative \emph{information} that
  represents probabilistic relational constraints on properties of
  objects in the world, augmenting the robot's sensory
  observations. For instance, a robot tasked with a search-and-rescue
  mission may be informed by the human that two victims are probably
  in the same room. An important question arises: how should we
  represent the robot's internal knowledge so that this information is
  correctly processed and combined with raw sensory information? In
  this paper, we provide an efficient belief state representation that
  dynamically selects an appropriate factoring, combining aspects of
  the belief when they are correlated through information and
  separating them when they are not. This strategy works in open
  domains, in which the set of possible objects is not known in
  advance, and provides significant improvements in inference time
  over a static factoring, leading to more efficient planning for
  complex partially observed tasks. We validate our approach
  experimentally in two open-domain planning problems: a 2D discrete
  gridworld task and a 3D continuous cooking task. A supplementary
  video can be found at \texttt{http://tinyurl.com/chitnis-iros-18}.
\end{abstract}

\section{Introduction}
As robots become increasingly adept at understanding and manipulating
the world around them, it becomes important to enable humans to
interact with them to convey goals, give advice, or ask questions. A
typical setting is a partially observed environment in which the robot
has uncertainty about its surroundings, but a human can give it
\emph{information} to help it act more intelligently. This information
could represent complex relationships among properties of objects in
the world, but the robot would be expected to use it as needed when
given a task or query. This setting motivates an important question:
what is the best way to represent the robot's internal knowledge so
that this information is correctly processed and combined with the
robot's own sensory observations? It is important for the chosen
representation to be able to accurately and efficiently answer queries
(i.e. do inference) that require it to draw on the given information.

\begin{figure}[t]
  \centering
    \noindent
    \includegraphics[width=\columnwidth]{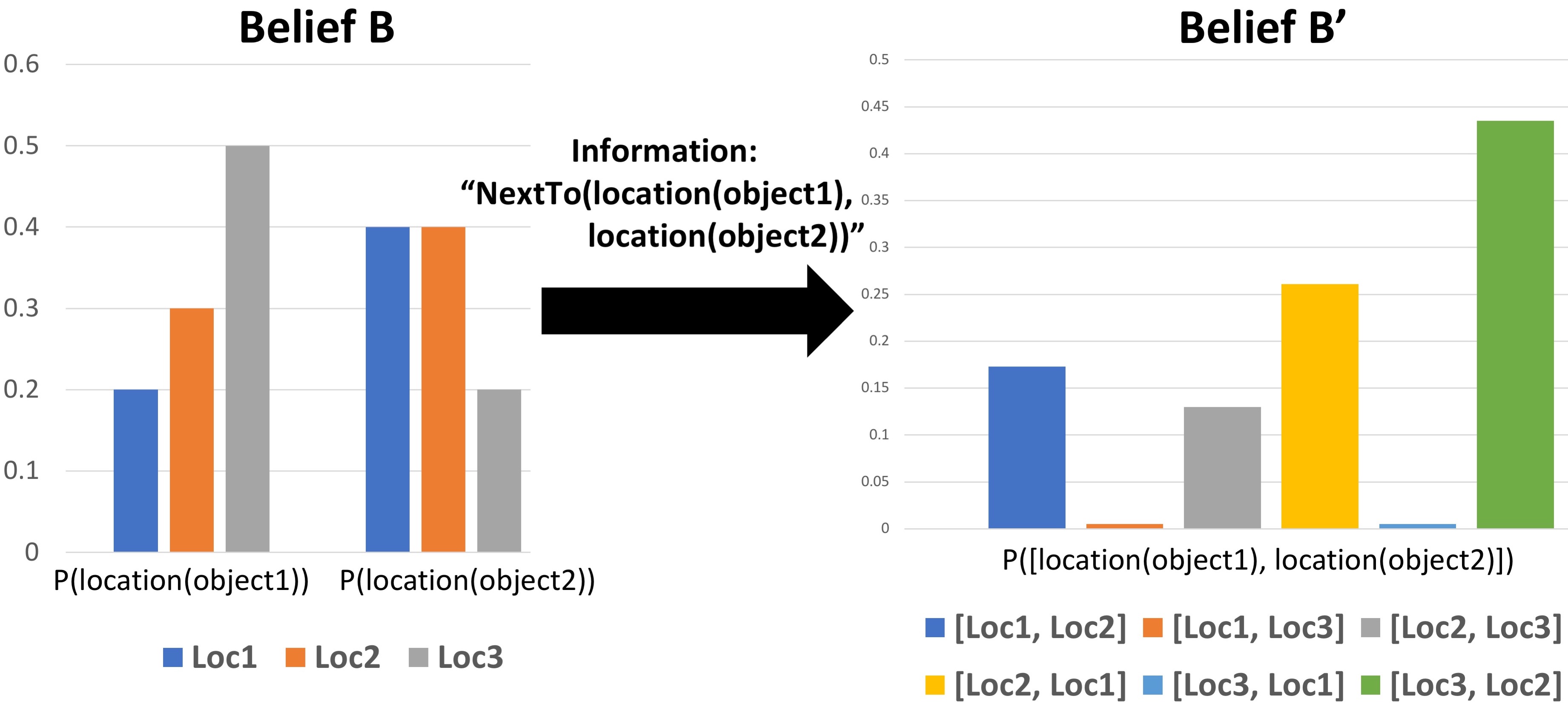}
    \caption{A schematic illustration of a dynamically factored
      belief. The initial belief $B$ tracks distributions over
      possible values for the locations of two objects. There are
      three locations in the world (not shown): Loc1 on the left, Loc2
      in the middle, and Loc3 on the right. When the agent is given
      information that the objects are next to each other, the belief
      is updated to produce $B'$. This is a new factoring in which the
      two old factors are joined into a single one, corresponding to a
      distribution over the joint location of both objects. Other
      factors (not shown) would not be affected.}
  \label{fig:teaser}
\end{figure}

We consider a specific class of open-domain planning problems in which
objects exist in the world, but the agent does not know the universe
of objects. We formalize our setting as a partially observable Markov
decision process in which the robot has two sources of (potentially
noisy) observations: its own perceptual capabilities, and
\emph{assertions} about the environment that simulate the human-provided
information and are expressed in formal language. These observations
represent constraints that hold with some probability and relate
properties of objects in the world.

In order to support inference in partially observed environments, one
typically maintains a belief state: a probability distribution over
the space of world states. Unfortunately, the full joint distribution
is usually intractable to work with. A popular alternative approach is
to represent a factored belief state, in which the world state is
decomposed into a set of features, each with a value. The factored
belief is then a mapping from every feature to a distribution over its
value.

We propose a method for efficient inference using a factored belief
state in the presence of potentially complicated assertions relating
multiple variables. The typical approach to using a factored belief
state involves committing to a (possibly domain-specific)
representational choice at the very
start~\cite{bonet2014belief,boyenkoller,sallans2000learning}, for
which can be difficult to fold in arbitrary relational constraints
without too much loss in accuracy. On the other hand, our work treats
a factored belief state as a fluid, dynamic data structure in which
the factoring itself is molded by the constraints, as suggested by
\figref{fig:teaser}. We call this a \emph{dynamically factored
  belief}.

For the class of open-domain planning problems we consider, we show
that a dynamically factored belief state representation provides
significant improvements in inference time over a fixed factoring. We
validate our approach experimentally in two open-domain planning
problems: a 2D discrete gridworld task and a 3D continuous cooking
task.

Visit \texttt{http://tinyurl.com/chitnis-iros-18} for a supplementary
video.

\section{Background}
\subsection{\pomdp s and Belief States}
We formalize agent-environment interaction in the presence of
uncertainty as a \emph{partially observable Markov decision process}
(\pomdp)~\cite{pomdp}. We consider a typical undiscounted setting
with: $\St$, the state space; $\A$, the action space; $\Obs$, the
observation space; $T(s, a, s') = P(s' \mid s, a)$, the transition
distribution with $s, s' \in \St, a \in \A$;
$O(s', a, o) = P(o \mid s', a)$, the observation model with
$s' \in \St, a \in \A, o \in \Obs$; and $R(s, a, s')$, the reward
function with $s, s' \in \St, a \in \A$. Some states in $\St$ are
\emph{terminal}, ending the episode.

At each timestep, the agent selects an action, causing 1) the hidden
state to change according to $T$, 2) the agent to receive a reward
according to $R$, and 3) the agent to receive an observation according
to $O$. The agent's objective is to maximize its overall expected
reward, $\mathbb{E}\left[\sum_{t}R(s_{t}, a_{t}, s_{t+1})\right]$. A
solution to a \pomdp\ is a policy that maps the history of
observations and actions to the next action to take, such that this
objective is optimized over the trajectory.

The sequence of states $s_{0}, s_{1}, ...$ is unobserved, so the agent
must instead maintain a \emph{belief state}: a probability
distribution over the space of world states. This belief is updated on
each timestep based on the received observation and taken action. The
exact belief update is
$B'(s') = \frac{1}{Z}\left[O(s', a, o) \sum_{s \in \St} T(s, a,
  s')B(s)\right],$ where $B$ and $B'$ are the old and new belief
states, $s' \in \St$ is a state, $a \in \A$ is the taken action,
$o \in \Obs$ is the received observation, and $Z$ is a normalizing
factor.

Representing the full belief exactly is prohibitively expensive
for even moderately-sized \pomdp s, so a typical alternative approach
is to use a \emph{factored} representation~\cite{boyenkoller}. Here, we assume that the
state can be decomposed into a set of features, each of which has a
value. The factored belief is then a mapping from every feature to a
distribution over its value. Typically, one chooses the features
carefully so that observations can be \emph{folded}, i.e.
incorporated, into the belief efficiently and without too much loss of
information. In other words, the chosen distributions are conjugate to
the most frequent kinds of observations. Most factored representations
are updated \emph{eagerly} (without explicitly remembering the actions and
observations) but approximately. On the other hand, a \emph{fully
  lazy} representation just appends $a$ and $o$ to a list at each
timestep. Though belief updates are trivial with this lazy
representation, inference can be very expensive. In our work, we will
give a representation that is sometimes eager and sometimes lazy,
based on how expensive it would be to perform an eager update.

Some popular approaches for generating policies in \pomdp s are online
planning~\cite{pomcp,despot,beliefspaceplanning} and finding a policy
offline with a point-based solver~\cite{sarsop,pointpomdp}. Our work
will use the more efficient but more approximate
\emph{determinize-and-replan} approach, which optimistically plans in
a determinized version of the environment, brought about by (for
instance) assuming that the maximum likelihood observation is always
obtained~\cite{plattmlobs,bsptamp}. The agent executes this plan and
replans any time it receives an observation contradicting the
optimistic assumptions made.

\subsection{Factor Graphs}
We will be viewing our approach from the perspective of factor graphs,
which we briefly describe in this section. We refer the reader to work
by Kschischang et al.~\cite{factorgraph} for a more thorough treatment. A \emph{factor
  graph} is a bipartite undirected probabilistic graphical model
containing two types of nodes: \emph{variables} and
\emph{factors}. Factor graphs provide a compact representation for the
factorization of a function. Suppose a function $f$ on $n$ variables
can be decomposed as
$f(X_{1}, X_{2}, ..., X_{n}) = \prod_{i=1}^{m}f_{i}(C_{i}),$ where
each $C_{i} \subseteq \{X_{1}, X_{2}, ..., X_{n}\}$ is a subset of the
variables. This decomposition corresponds to a factor graph in which
the variables are the $X_{j}$, the factors are the $f_{i}$, and there
is an edge between any $f_{i}$ and $X_{j}$ for which
$X_{j} \in C_{i}$, i.e. $f_{i}$ is a function of $X_{j}$.

This representation affords efficient marginal inference, which is the
process of computing the marginal distribution of a variable, possibly
conditioned on the values of some other variables. Message-passing
algorithms such as the sum-product algorithm typically compute
marginals using dynamic programming to recursively send messages
between neighboring nodes. The sum-product algorithm is also commonly
referred to as \emph{belief propagation}.


\section{Related Work}
We focus on the setting of \emph{information-giving} for open-domain
human-robot collaboration. Information-giving was first explored
algorithmically by McCarthy~\cite{programswithcommonsense} in a
seminal 1959 paper on advice-takers. Much work in open-domain
collaboration focuses on the robot understanding goals given by the
human~\cite{talamadupula1}, whereas we focus on understanding
information given by the human, as in work on
advice-giving~\cite{odom2015learning} and commonsense
reasoning~\cite{corpp}.

\subsection{Adaptive Belief Representations}
Our work explores belief state representations that adapt to the
structure of the observations received by the robot. The work perhaps
most similar to ours is that of Lison et
al.~\cite{beliefsituationaware}, who acknowledge the importance of
information fusion and abstraction in uncertain environments. Building
on Markov Logic Networks~\cite{mln}, they describe a method for belief
refinement that 1) groups percepts likely to be referring to the same
object, 2) fuses information about objects based on these percept
groups, and 3) dynamically evolves the belief over time by combining
it with those in the past and future. They focus on beliefs about each
object in the environment, whereas our work focuses on combining
information about multiple objects, based on the structure of the
observations.

The notion of adaptive belief representations has also been explored
in domains outside robotics. For instance, Sleep~\cite{adaptivegrid}
applies this idea to an acoustic target-tracking setting. The belief
representation, which tracks potential locations of targets, can
expand to store additional information about the targets that may be
important for locating them, such as their acoustic power. The belief
can also contract to remove information that is deemed no longer
necessary. It would be difficult to update this factoring with
information linking multiple targets, whereas our method is
well-suited to incorporating such relational constraints.

\subsection{Factored Belief Representations for \pomdp s}
The more general problem of finding efficient belief representations
for \pomdp s is very well-studied. Boyen and Koller~\cite{boyenkoller}
were the first to provide a tractable method for belief propagation
and inference in a hidden Markov model or dynamic Bayesian
network. Their basic strategy is to first pick a computationally
tractable approximate belief representation (such as a factored one),
then after a belief update, fold the newly obtained belief into the
chosen approximate representation. This technique is a specific
application of the more general principle of \emph{assumed density
  filtering}~\cite{maybeck1982stochastic}. Although it seems that the
approximation error will build up and propagate, the authors show that
actually, the error remains bounded under reasonable assumptions. Our
work adopts a fluid notion of a factored belief representation, not
committing to one factoring.

Bonet and Geffner~\cite{bonet2014belief} introduce the idea of beam tracking for belief
tracking in a \pomdp. Their method leverages the fact that a problem
can be decomposed into projected subproblems, allowing the joint
distribution over state variables to be represented as a product over
factors. Then, the authors introduce an alternative decomposition over
beams, subsets of variables that are causally relevant to each
observable variable. This work shares with ours the idea of having the
structure of the belief representation not be fixed ahead of time. A
difference, however, is that the decomposition used in beam tracking
is informed by the structure of the underlying model, while ours is
informed by the structure of the observations and can thus efficiently
incorporate arbitrary relational constraints on the world state.


\section{Formal Problem Setting}
In this section, we formalize our problem setting as a \pomdp, for
which we will show that a dynamically factored belief is a good
representation for the agent's belief state.

\emph{Assumptions.} We will assume deterministic transitions and a uniform
observation model over all valid observations, in order to make
progress on this difficult problem.

\subsection{Planning Problem Class}
We first describe the underlying class of planning problems. We
consider \emph{open domains}, in which the world contains a
(finite or infinite) universe of objects, but the agent does not know this
universe. Planning in open domains is significantly more complex than
planning in settings where the universe of objects is known in
advance.

The class of \emph{open-domain planning problems} $\Pi$ contains tuples
$\langle \T, \P, \Obj, \V, \F, \U, \I, \G \rangle$:
\begin{tightlist}
\item $\T$ is a known set of object \emph{types}, such as locations or
  movables. For some types, the set of objects may be
  known to the agent in advance; for others, it may not.
\item $\P$ is a known set of \emph{properties} (such as color,
  size, pose, or contents) for each type from $\T$. Each property has
  an associated (possibly infinite) domain.
\item $\Obj$ is a (possibly partially) unknown set of \emph{objects},
  each of a type from $\T$. This set $\Obj$ can be finite or infinite.
\item $\V$ is the set of \emph{state variables} resulting from
  applying each property in $\P$ to every object in $\Obj$ of the
  corresponding type. Each variable has a domain based on the
  property and can be either continuous or discrete.
\item $\F$ is a set of \emph{fluents} or \emph{constraints},
  Boolean-valued expressions given by a predicate applied to state
  variables and (possibly) values in their domains. Examples:
  \emph{Equal(size(obj1), 6)}; \emph{Different(color(obj2),
    color(obj3))}.
\item $\U$ is a set of object-parametrized \emph{operators} that
  represent ways the agent can affect its environment. Each has
  preconditions (partial assignment of values to $\V$ that must hold for
  it to be legal), effects (partial assignment of values to $\V$ that
  holds after it is performed), and a cost.
\item $\I$ is an assignment of values to $\V$ defining the \emph{initial state}.
\item $\G$ is a partial assignment of values to $\V$ defining the \emph{goal}.
\end{tightlist}

A solution to a problem in $\Pi$ is a minimum-cost sequence of
parametrized operators $u_{1}, ..., u_{n} \in \U$ (a \emph{plan}) such
that starting with $\I$ and applying the $u_{i}$ sequentially
satisfies operator preconditions and causes the partial assignment
$\G$ to hold. Variables in $\V$ that are not in $\G$ may have any
value.

\subsection{\pomdp\ Formulation}
With $\Pi$ defined, we are ready to formulate our setting as a
\pomdp. Let $\langle \T, \P, \Obj, \V, \F, \U, \I, \G \rangle$ be an
open-domain planning problem from $\Pi$. Define the \pomdp:
\begin{tightlist}
\item $\St$ (the state space) is the space of all possible assignments
  of values to $\V$. A state is, thus, an assignment of a value to
  each variable in $\V$. Note that $\I$ is a state.
\item $\A$ (the action space) is $\U$.
\item $\Obs$ (the observation space) is the space of all (potentially
  noisy) observations: we define an observation as a set of pairs
  $\langle f, p \rangle$, where $f \in \F$ and $p \in (0,
  1]$. Intuitively, the interpretation is that every fluent (constraint) $f$ in this
  set holds with probability $p$ in the current state. We assume each
  observation $o$ comes from either 1) the robot's own perceptual
  capabilities or 2) an \emph{assertion} about the environment, which
  simulates human-provided information. For each fluent $f$ in $o$,
  the corresponding $p$ is a measure of confidence in $f$ holding within the
  current state. It can be based on the quality of a sensor or on the
  human's certainty about the veracity of their assertion. Because
  $\Obj$ is unknown, a fluent may contain state variables referencing
  objects the agent has not encountered before.
\item $T(s, a, s')$ (the transition distribution) is 1 if $s$
  satisfies $a$'s preconditions and $s'$ its effects; 0 otherwise.
\item $O(s', a, o)$ (the observation model) is a uniform distribution
  over all valid observations.
\item $R(s', a)$ (the reward function) is the negative cost of $a$.
\item $s \in \St$ is \emph{terminal}, ending the episode, if $\G$ holds in $s$.
\end{tightlist}

A solution to this \pomdp\ is a policy that maps the history of
observations and actions to the next action to take, such that the sum
of expected rewards is maximized: observe that this corresponds to
solving $\Pi$. Sources of uncertainty in this formulation are the open
domain and the noisy observations.

Next, we present the \emph{dynamically factored belief} as a good
belief state representation for this \pomdp. Though this
representation does not depend on the presence of assertions or an
open domain, we present it within this context because it best
motivates the approach and manifests its strengths.

\section{Dynamically Factored Belief}
\subsection{Overview}
In trying to find a suitable belief state representation for the
\pomdp\ defined in the previous section, we must be cognizant of the
fact that the agent does not know $\Obj$, the complete set of objects
in the world. A natural representation to use might be a factored one
over each state variable in $\V$, but unfortunately, if $\Obj$ is unknown
then so is $\V$. Furthermore, fluents in the observations may be
complicated expressions involving multiple state variables; we would
like our representation to be able to incorporate these observations.

\begin{figure}[t]
  \vspace{0.6em}
  \centering
    \noindent
    \includegraphics[width=\columnwidth]{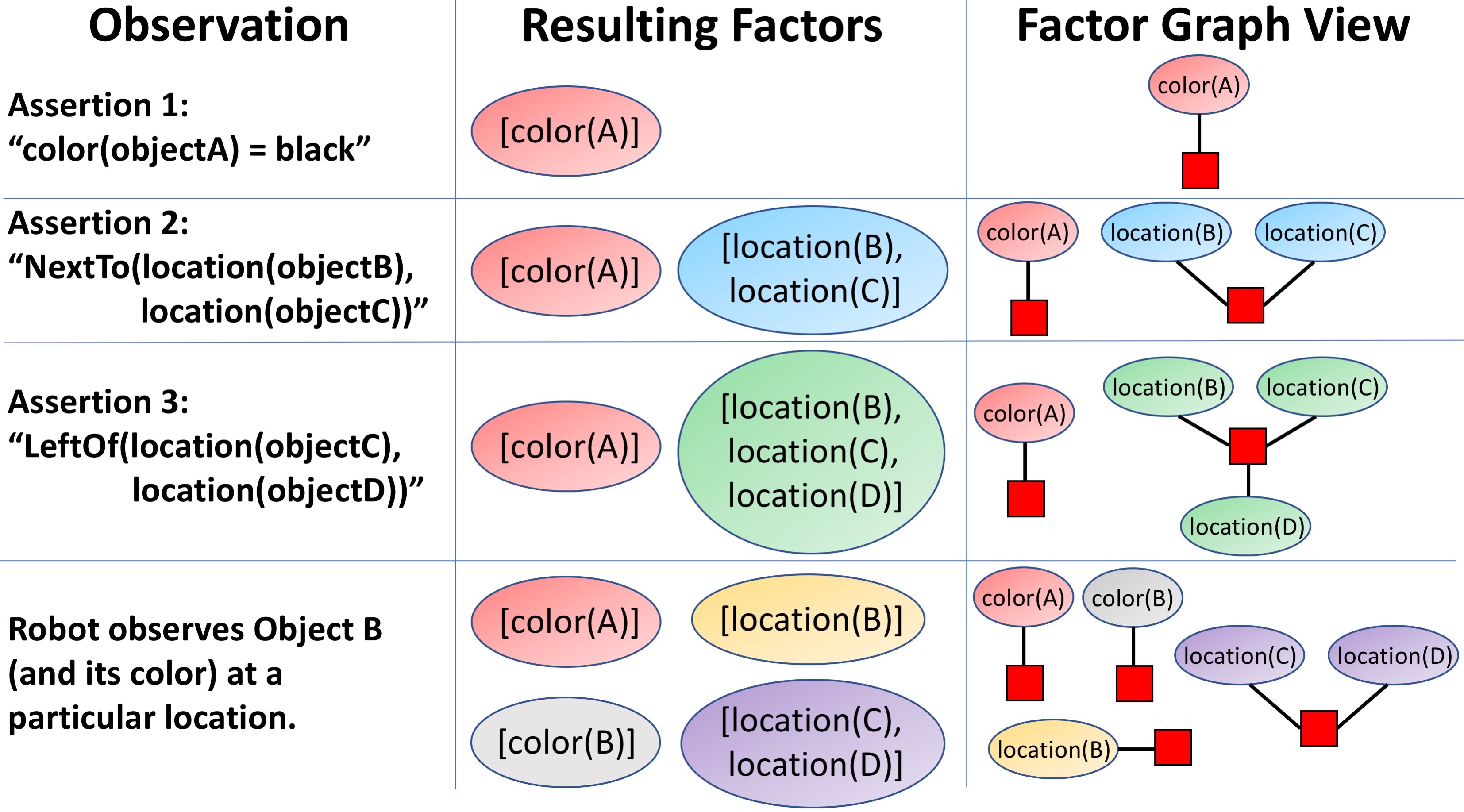}
    \caption{Example of dynamic factoring given four sequential
      noiseless observations in a setting with one object type and two
      properties: \emph{color} and \emph{location}. Each factor maps
      to a distribution over values (not shown). Initially, there are
      no factors. \emph{Row 1:} The agent receives an assertion
      containing one state variable, \emph{color(A)}; a singleton
      factor for this variable is introduced. \emph{Row 2:} The
      assertion contains two state variables, so a joint factor is
      introduced. \emph{Row 3:} The assertion contains a new state
      variable, \emph{location(D)}, so a factor is introduced for it,
      then joined with the [\emph{location(B), location(C)}]
      factor. \emph{Row 4:} The agent observes the color and location
      of Object B. The joint distribution over the locations of B, C,
      and D is now uniform across the first dimension, so the factor
      gets split into two. This type of splitting implies that factors
      do not necessarily get bigger each time a new observation
      arrives. In this figure, all observations are noiseless (all
      $p=1$) for clarity of presentation. The factor graph is always a
      collection of independent subgraphs, where each subgraph
      contains exactly one factor.}
  \label{fig:dfboverview}
\end{figure}

We note the following. First, we do not need to know the full set of
state variables in order to maintain a (partial) factored belief
representation. Second, the choice of factors should be dynamic and
influenced by the observations, allowing us to gracefully cope with
complicated assertions. The intuition is that a constraint linking two
state variables would not be foldable into a factored representation
over each state variable, so the representation should be
modified. Building on these ideas, we present the following
definition.

\begin{defn}
  A \emph{dynamically factored belief state representation} has two components:
  \begin{tightlist}
  \item A factored representation for which each factor is a
    \emph{list} of one or more state variables from $\V$. The factors
    partition the set of state variables that the agent knows about so
    far, either from prior knowledge or from a received
    observation. Each factor maps to a joint probability distribution
    over possible values for all its variables.
  \item A database called ComplexFluents.
  \end{tightlist}
\end{defn}

The belief is initialized to have a singleton factor for each state
variable in $\V$ associated with an object in $\Obj$ known a priori to
exist, mapping to any distribution (e.g. a prior).

Each time an observation $o \in \Obs$ is received, the belief is
updated as follows with every constituent fluent $f$: all factors
containing state variables mentioned by $f$ are introduced (if they
are not represented yet) and joined, so long as the resulting joint
would not be too big. If it would be too big, then the fluent is
lazily placed into ComplexFluents and considered only at query
time. Furthermore, factors are regularly split up for efficiency,
implying that a belief update could potentially compress the
representation. This is shown in \figref{fig:dfboverview} and
described in detail in the next section. Newly introduced variables
can map to any distribution, such as a uniform one, or one calculated
from some prior knowledge.

The factors partition the set of state variables that the agent knows
about so far, meaning no state variable can ever be present in more
than one factor. Enforcing this invariant greatly simplifies
inference, as can be understood from a factor graph perspective. A
dynamically factored belief maintains a factor graph with variable
nodes $V$ and factor nodes $F$, where the $V$ are the state variables
in $\V$ that the agent knows about so far, and the $F$ are the
factors. Each node $F$ connects to all the $V$ that are present in the
corresponding list (and thus comprise that factor's joint
distribution). Because the factors partition the variables, this
factor graph is a collection of independent subgraphs, where each
subgraph contains exactly one node from $F$ connected to one or more
nodes from $V$. See \figref{fig:dfboverview} for an example. Fluents
placed lazily into ComplexFluents are not part of this factor
graph. The structure of this factor graph is constantly changing as
new objects are discovered and observations are received, but it will
always be a collection of independent subgraphs. Thus, there is no
possibility of ending up with a cyclic factor graph, which would
typically necessitate either approximate inference or expensive
exact inference.

\emph{Note.} All fluents, complex or not, can be represented by a
factor in some more complex factor graph. One could imagine a
different algorithm that avoids constructing the joint of all
variables mentioned by a fluent, but instead creates a new factor for the
fluent while leaving other factors untouched. This would likely create
cycles, but cycles could be collapsed into joint distributions over
all constituent nodes. Inference would then be done using
message-passing. In contrast, our method eagerly incorporates fluents
in a way that makes inference fast and cycles impossible. Furthermore,
our method can very quickly answer queries about a marginal on a
subset of any factor, as we will see.

\subsection{Belief Update Algorithm}
\begin{algorithm}[t]
  \SetAlgoLined
  \SetAlgoNoEnd
  \DontPrintSemicolon
  \SetKwFunction{algo}{algo}\SetKwFunction{proc}{proc}
  \SetKwProg{myalg}{Algorithm}{}{}
  \SetKwProg{myproc}{Subroutine}{}{}
  \SetKw{Continue}{continue}
  \SetKw{Return}{return}
  \myalg{Dynamically Factored Belief Update}{
    \nl $B \gets$ InitializeFactoredBeliefMap()\;
    \nl ComplexFluents $\gets$ set(\{\})\;
    \myproc{\textsc{BeliefUpdate}(observation, action)}{
        \nl \For{each $\langle f, p \rangle$ in observation}{
          \nl \For{each stateVar mentioned by $f$}{
            \nl \If{!$B$.Contains(stateVar)}{
              \nl $B$.Add([stateVar], defaultDist())\;
            }
          }
          \nl \eIf{joint would be too big}{
            \nl ComplexFluents.Add($f$)\;
            }{
          \nl $B$.JoinFactorsAndUpdate($f, p$)\;}
        }
        \nl $B$.UpdateWithAction(action)\;
        \tcp{\footnotesize Keep representation compact.}
        \nl \For{each factor in $B$.Factors}{
          \nl $B$.TrySplit(factor, $\epsilon$)\;
        }
        }}\;
\caption{Dynamically factored belief update.}
\label{alg:dynamicbeliefupdate}
\end{algorithm}

\algref{alg:dynamicbeliefupdate} gives pseudocode for updating a
dynamically factored belief. The belief is initialized to contain a
singleton factor for each state variable in $\V$ associated with an
object in $\Obj$ known a priori to exist. The \textsc{BeliefUpdate} subroutine is called at
each timestep, after the agent takes an action and receives an
observation (a set of fluents, each with a corresponding probability
of holding in the world).

Line 7 decides whether joining all factors containing a state variable
mentioned by the fluent would result in a joint that is too ``big'' (expensive to compute or represent). If so, the fluent is
lazily stored in the database ComplexFluents and considered only at
query time. An implementation of this test could take the product of
the factor sizes and check whether it is above a certain threshold.


\begin{algorithm}[t]
  \SetAlgoLined
  \SetAlgoNoEnd
  \DontPrintSemicolon
  \SetKwFunction{algo}{algo}\SetKwFunction{proc}{proc}
  \SetKwProg{myalg}{Algorithm}{}{}
  \SetKwProg{myproc}{Subroutine}{}{}
  \SetKw{Continue}{continue}
  \SetKw{Return}{return}
  \myproc{\textsc{JoinFactorsAndUpdate}($B, f, p$)}{
      \nl joint $\gets$ (join factors $F$ containing variables in $f$)\;
      \nl $m \gets$ (total prob. of values inconsistent with $f$)\;
      \nl \For{each value in joint}{
        \nl \If{value is inconsistent with $f$}{
          \nl joint.RescaleProbBy$\left(\text{value}, \frac{(1-p)(1-m)}{pm}\right)$\;
        }
      }
      \nl joint.Normalize()\;
      \nl Add joint to $B$ and remove all $F$ from $B$.\;
    }\;

    \myproc{\textsc{TrySplit}($B$, factor, $\epsilon$)}{
      \nl \For{each state variable $V$ in factor}{
        \nl reconstructed $\gets$ Join($B$[V], $B$[factor $\setminus$ \{V\}])\;
        \nl \If{$\js{B[\text{factor}]}{\text{reconstructed}} < \epsilon$}{
          \nl Split $B$[factor] into $B$[V], $B$[factor $\setminus$ \{V\}].\;
        }
      }
    }\;
  \caption{Subroutines used in \algref{alg:dynamicbeliefupdate}.}
\label{alg:subroutines}
\end{algorithm}

\begin{figure}[t]
  \vspace{0.6em}
  \centering
    \noindent
    \includegraphics[width=0.45\textwidth]{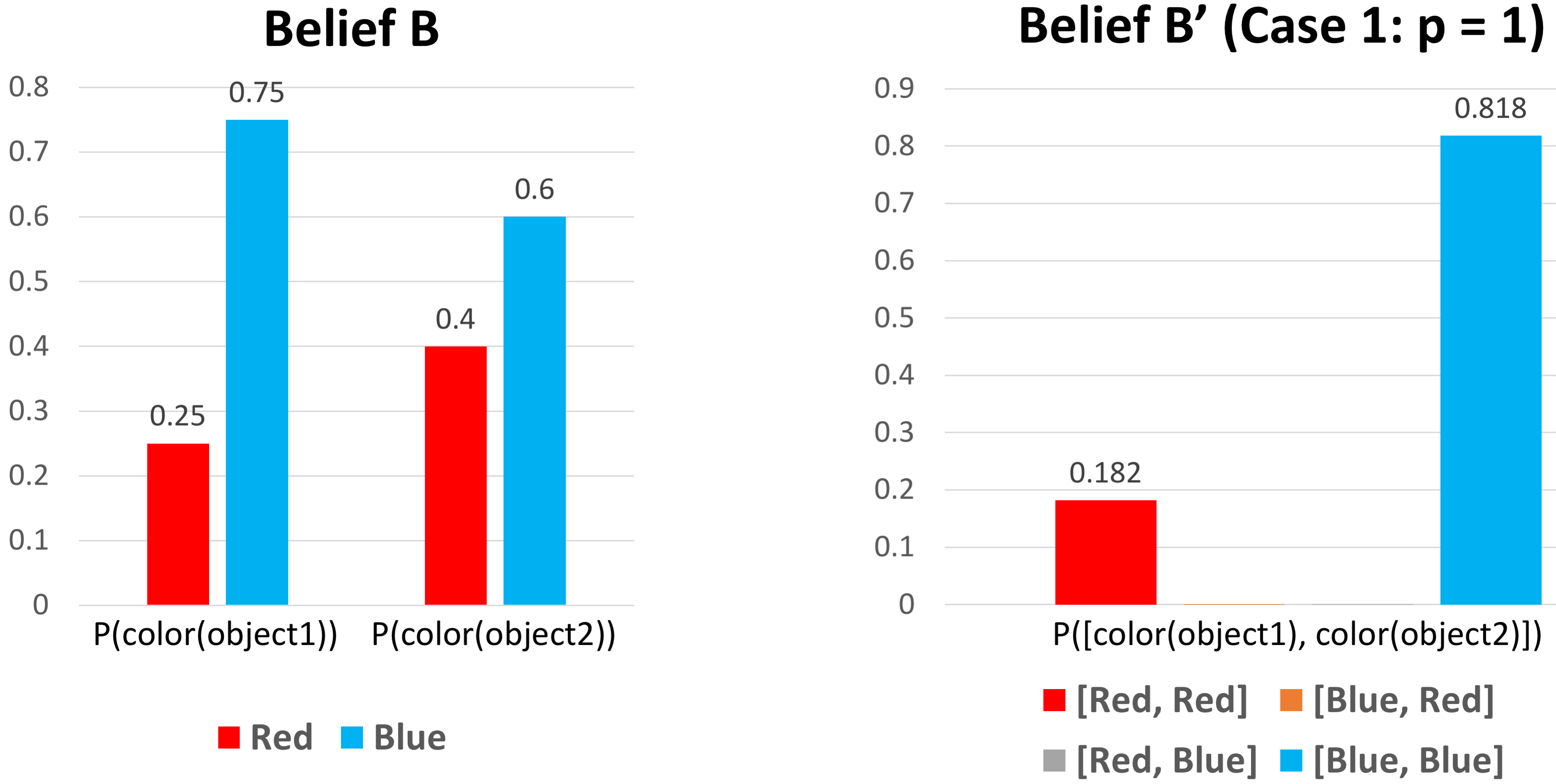}

    \vspace{0.7em}

    \includegraphics[width=0.45\textwidth]{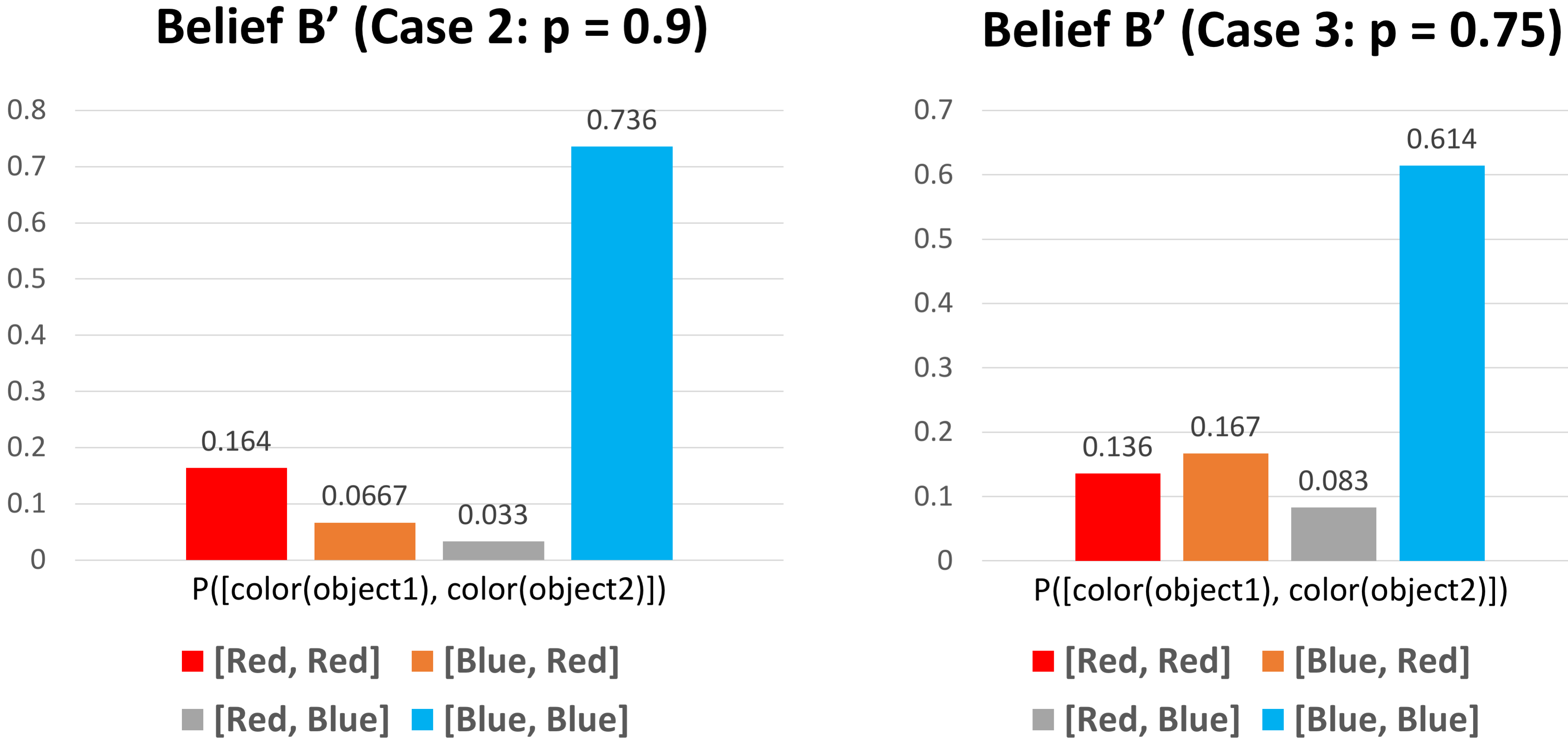}
    \caption{Posteriors from running \textsc{JoinFactorsAndUpdate}
      with fluent \emph{Same(color(object1), color(object2))}, for
      three values of $p$.}
  \label{fig:joinfactorsandupdate}
\end{figure}

\textbf{Joining Factors} \ The call in Line 9 to
\textsc{JoinFactorsAndUpdate} creates a new factor containing all
state variables that are mentioned by the fluent $f$ (if such a factor
is not already present), then maps this factor to a joint distribution
for which $f$ holds with probability $p$. To accomplish this, we build
the joint then rescale the probabilities such that $p$ mass goes to
the joint values which are consistent with $f$, and the remaining
$1-p$ mass goes to those which are inconsistent. This can be done
using Jeffrey's rule~\cite{jeffreysrule}: rescale the probability of
all values inconsistent with $f$ by $\frac{(1-p)(1-m)}{pm}$, where $m$
is the total probability of these inconsistent values, then
normalize. When $p = 1$, this algorithm just filters out joint values
inconsistent with $f$, as expected. See \algref{alg:subroutines} for
pseudocode and \figref{fig:joinfactorsandupdate} for an example. If
distributions are continuous, we perform these operations implicitly
using rejection sampling at query time.

\textbf{Splitting Factors} \ The call in Line 12 to \textsc{TrySplit}
attempts to split up factors to maintain a compact representation. In
practice, we accomplish this by checking whether each state variable
in the factor can be marginalized out. Of course, it is unlikely that
such marginalization can ever be done in a lossless manner, as this
would require the joint to be exactly decomposable into a product
involving this marginal. Instead, we perform marginalization whenever
the reconstruction error is less than a hyperparameter $\epsilon$. We
measure this reconstruction error as the Jensen-Shannon divergence
$D_{JS}$ between the true joint and the approximate joint
reconstructed from the attempted decomposition.

\begin{figure}[t]
  \centering
    \noindent
    \includegraphics[width=0.45\textwidth]{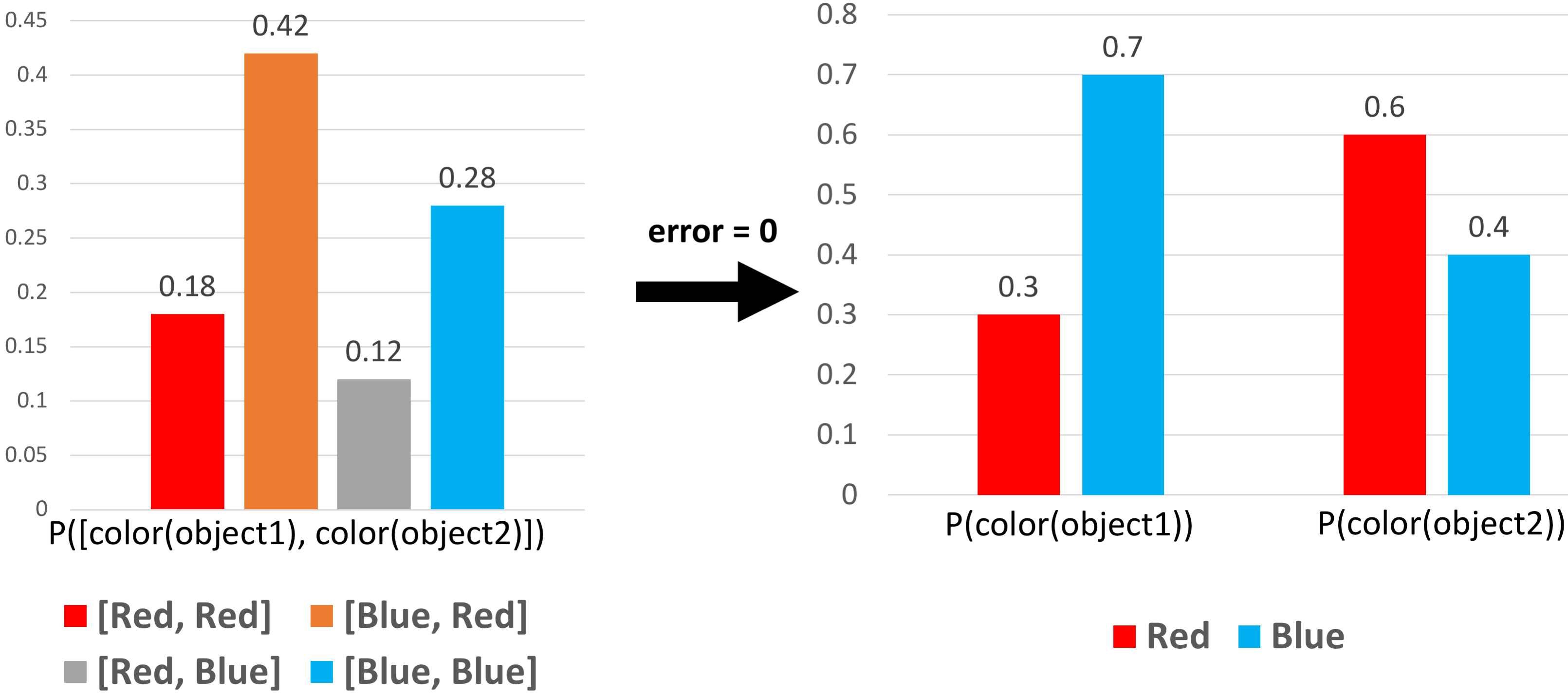}

    \vspace{0.7em}

    \includegraphics[width=0.45\textwidth]{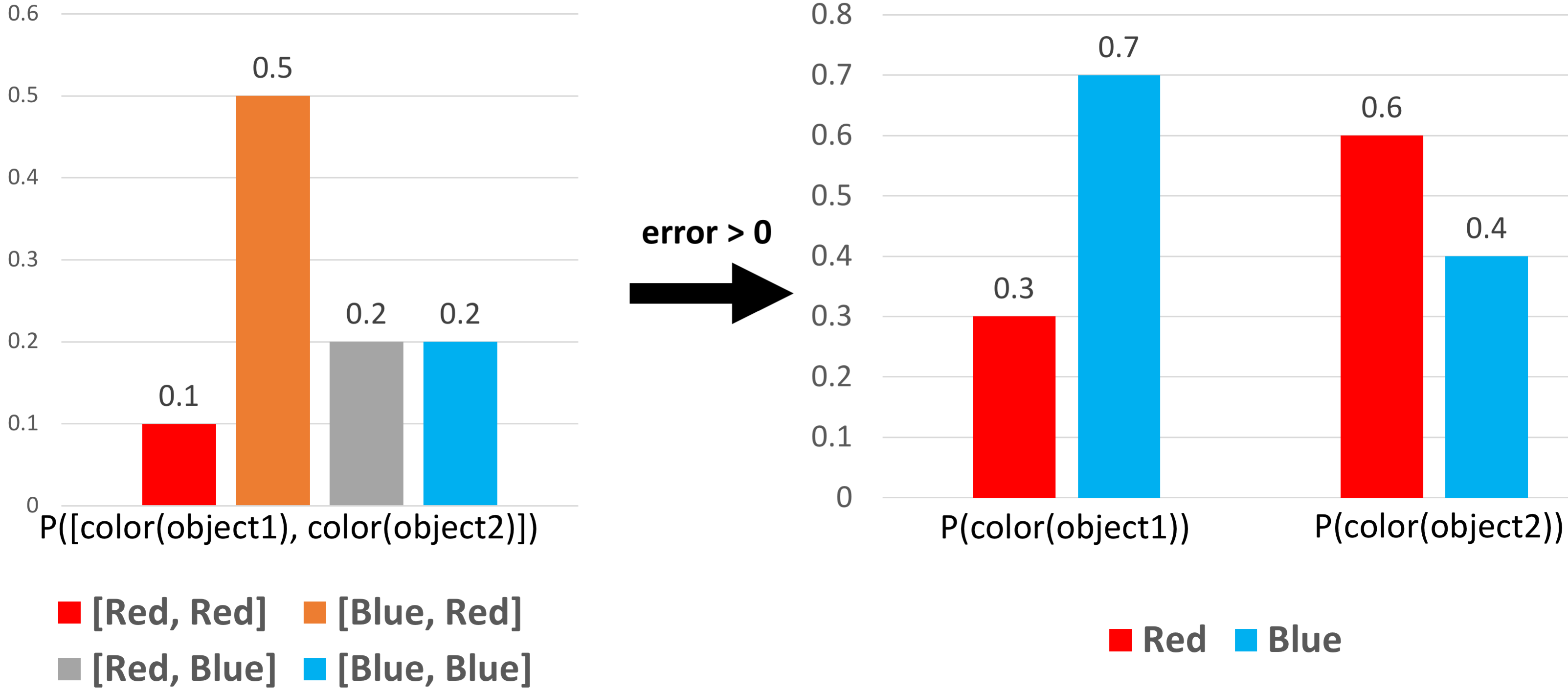}
    \caption{Two examples of trying to split up a factor having two
      variables. \emph{Top:} The variables are independent; no error
      is incurred. \emph{Bottom:} Error is incurred based on the
      Jensen-Shannon divergence between the joint and the product of
      the marginals ($\sim$0.01 here).}
  \label{fig:trysplit}
\end{figure}

Let $P$ and $Q$ be arbitrary probability distributions. The
Jensen-Shannon divergence is a smooth, symmetric, bounded measure of
similarity between $P$ and $Q$. It is based on the Kullback-Leibler
(KL) divergence, defined as
$\kl{P}{Q} = -\sum_{i}P(i) \log \frac{Q(i)}{P(i)},$ where the
summation can be replaced by integration for continuous
distributions. Then letting $A = \frac{1}{2}(P+Q)$, the Jensen-Shannon
divergence is defined as
$\js{P}{Q} = \frac{1}{2} \kl{P}{A} + \frac{1}{2} \kl {Q}{A}.$ Assuming
the natural logarithm is used, the bound
$0 \leq \js{P}{Q} \leq \log 2$ always holds. Since $D_{JS}$ is
bounded, it is reasonable to use $\epsilon$ as a fixed threshold on
the reconstruction error to decide whether to do
marginalization. Varying $\epsilon$ lets the designer trade off
between compactness of the belief and accuracy of inference. See
\algref{alg:subroutines} for pseudocode and \figref{fig:trysplit} for
an example.

\subsection{Inference}
Dynamically factored beliefs can handle two types of queries: 1) a
marginal on any state variable, or on a set of state variables that is
a subset of some factor, and 2) a sample from the full joint of
current factors, which gives a world state consistent with the agent's
current knowledge.

To answer queries of type 1), we observe that if a state variable or set of state
variables is a subset of some factor, then our representation already
stores a joint distribution over values for those
variables. Any query about them can be answered
using this joint, e.g. by sampling. We do not have
to worry about violating the constraints in ComplexFluents because
each of those could only ever constrain a set of state variables that
spans multiple factors.

To answer queries of type 2), we must draw from the distribution implicitly encoded by
all factors together, which produces an assignment that maps every
currently known state variable to a value. One can
treat this as a constraint satisfaction problem~\cite{csp} where
the constraints are the fluents in ComplexFluents (all other fluents
have already been folded eagerly into the belief), and apply standard
solving techniques. Our experiments solve it incrementally using a backtracking
approach; see \algref{alg:samplestate} for pseudocode.

\begin{algorithm}[t]
  \SetAlgoLined
  \SetAlgoNoEnd
  \DontPrintSemicolon
  \SetKwFunction{algo}{algo}\SetKwFunction{proc}{proc}
  \SetKwProg{myalg}{Algorithm}{}{}
  \SetKwProg{myproc}{Subroutine}{}{}
  \SetKw{Continue}{continue}
  \SetKw{Return}{return}
  \myalg{\textsc{SampleState}($B$, ComplexFluents)}{
      \nl state $\gets$ map(each state variable in $B \to$ NULL)\;
      \nl curIndex $\gets$ 0\;
      \nl \While{curIndex $<$ number of factors}{
        \nl factor $\gets$ $B$.Factors[curIndex]\;
        \nl \If{sampling limit reached}{
          \nl \For{each stateVar in factor}{
            \nl state[stateVar] $\gets$ NULL\;}
          \nl curIndex $\gets$ curIndex $-$ 1\;
          \Continue}
        \nl values $\gets B$[factor].Sample()\;
        \nl \For{stateVar, value in zip(factor, values)}{
          \nl state[stateVar] $\gets$ value\;}
        \nl \If{any $f$ in ComplexFluents cannot hold}{\Continue}
        \nl curIndex $\gets$ curIndex + 1\;
      }
      \Return state
    }\;
\caption{An incremental algorithm for sampling a world state
  consistent with all observations, using a dynamically factored
  belief $B$. The returned state is an assignment of currently known state
  variables to values.}
\label{alg:samplestate}
\end{algorithm}

\section{Experiments}
We evaluate the performance of our approach on the cooking task, a
planning problem from $\Pi$. The robot is tasked with gathering
ingredients and using them to cook a meal. There are three object
types: $\T = \{$\emph{locations}, \emph{vegetables},
\emph{seasonings}$\}$. The term \emph{ingredients} refers to
vegetables and seasonings together. Each object type has one
property. Locations have a \emph{contents} property, which is one of
``vegetable,'' ``seasoning,'' or ``empty.'' Ingredients have a
\emph{position} property, which could be continuous- or
discrete-valued. Initially, the robot knows the set of locations but
not the set of ingredients. There is a pot at a fixed, known position.

When any vegetable is placed into the pot, it transitions to a
\emph{cooking} state; 5 timesteps later, it transitions to a
\emph{cooked} state. The goal is to have all ingredients in the pot
and all vegetables cooked. However, the robot is penalized heavily for
placing any seasoning into the pot too early, before all vegetables
have been cooked. To achieve the goal, the robot must learn the
positions of all ingredients, either by doing observations or by
learning about them from assertions.

The operators (actions) $\U$ are:
\begin{tightlist}
\item \textsc{Observe(location)}: Moves and observes the ingredient(s)
  at a location. Cost: 5.
\item \textsc{Pick(position)}: Moves and picks at a continuous- or
  discrete-valued position (domain-dependent). The robot can
  hold up to 10 ingredients at once. Cost: 20.
\item \textsc{PlaceInPot()}: Places all $n$ held ingredients into the
  pot. Vegetables in the pot are either \emph{cooking} or, 5 timesteps
  later, \emph{cooked}. Cost: 100 + 50$n$, plus an additional 1000 if
  a seasoning is placed in before all vegetables are cooked.
\item \textsc{No-op()}: Takes no action. Cost: 0.
\end{tightlist}
There is also a living cost of 10 per timestep.

The state variables $\V$ comprise each location's contents and each
ingredient's position. The world state contains an assignment of these
variables to values. It also tracks which ingredients are held by the
robot and the pot, and the current robot pose; these are all assumed
to be known and thus do not need to be tracked by the belief state.

\begin{figure}[t]
  \vspace{0.6em}
  \centering
    \noindent
    \includegraphics[width=\columnwidth]{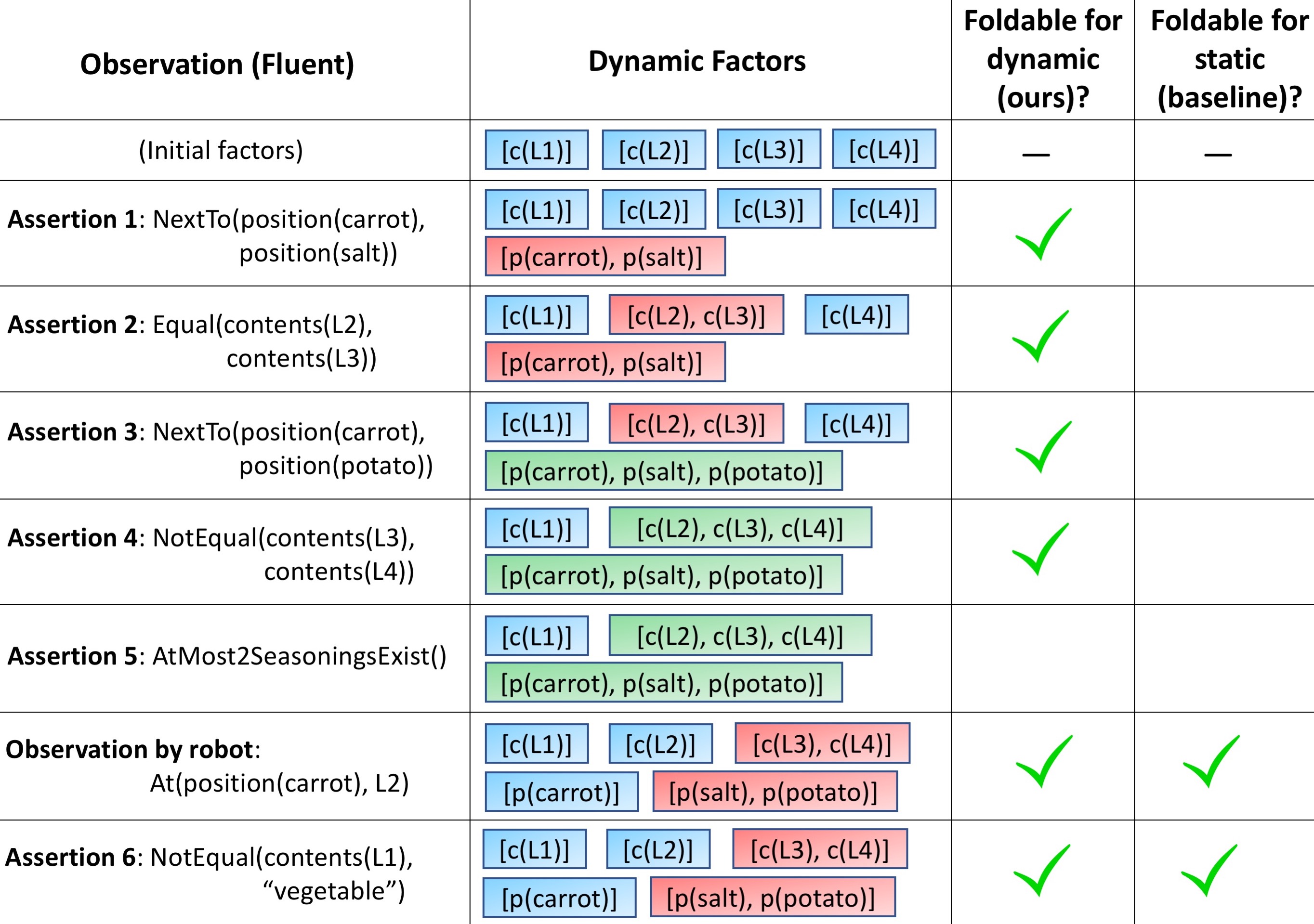}
    \caption{An illustration of the types of assertions we use, with a
      simplified execution of the gridworld cooking task. There are
      four locations: L1, L2, L3, L4. Factors are color-coded based on
      the number of state variables. c($\cdot$) means
      \emph{contents($\cdot$)}, p($\cdot$) means
      \emph{position($\cdot$)}. Initially, there is a singleton factor
      for each location's contents. The last two columns tell whether
      the fluent is foldable into a dynamic factoring (our approach),
      and into a static factoring that tracks the potential contents
      of each location (baseline). Unfoldable fluents get placed into
      ComplexFluents, slowing down inference. Observe that singleton
      factors go in and out of joints.}
  \label{fig:assertions}
\end{figure}

\emph{Assertions.} \figref{fig:assertions} shows the types of
assertions we use. At each timestep, we sample an assertion uniformly
at random from all valid ones, following our observation model, and
give it to the robot. The information could be redundant.

\emph{Baseline.} We test against a baseline belief representation that
simulates prior work on factored representations, which typically
commit to a representational choice at the start. This baseline, which
we call a \emph{statically factored belief} or \emph{static
  factoring}, represents the agent's belief as a distribution over the
potential contents of each location. The factoring does not change
based on observations, or as ingredients are discovered. We choose
this factoring as our baseline because initially, the robot only knows
the set of locations, not the set of ingredients. Thus, this factoring
is the most reasonable choice for a static representation that is
chosen at the start and held fixed throughout execution. Any fluent
that cannot fold into this static factoring is lazily placed into
ComplexFluents and considered only at query time.

\textbf{Domain 1: Discrete 2D Gridworld Cooking Task} \ Our first
experimental domain is the cooking task in a 2D gridworld. Locations
are organized in a 2D grid, and the robot is in exactly one at any
time. Both \textsc{Observe} and \textsc{Pick} actions are performed on
single grid locations. Each location is initialized to contain either
a single ingredient or nothing, so the \emph{position} property of
each ingredient is a discrete value corresponding to a location. We
solve the task using the determinize-and-replan approach to belief
space planning, with A* as the planner. The task is computationally
challenging, so we impose a 60-second timeout per episode.

\textbf{Domain 2: Continuous 3D Cooking Task} \ Our second
experimental domain is the cooking task in a pybullet
simulation~\cite{pybullet}. Here, the \emph{position} property of each
ingredient is a continuous value: each can be placed anywhere on the
surface of one of four tables. The robot can \textsc{Observe} any
table and \textsc{Pick} at any position along a table surface, so the
action space is continuous. Locations are given by a grid
discretization of the environment geometry. As ingredients get
discovered, a dynamically factored belief adapts to track continuous
distributions over their positions. Again, we use the
determinize-and-replan method and a 60-second timeout.

\begin{table}[t]
  \vspace{0.6em}
  \centering
  \resizebox{\columnwidth}{!}{
  \tabcolsep=0.08cm{
  \begin{tabular}{c|c|c|c|c}
    \toprule[1.5pt]
    \textbf{Setting} & \textbf{System} & \textbf{\% Solved} & \textbf{Bel. Upd. Time} & \textbf{Queries / Second}\\
    \midrule[2pt]
    Dom. 1, 4x4, 6ing & S (baseline) & 67 & 0.01 & 0.11\\
    \midrule
    Dom. 1, 4x4, 6ing & D (ours) & \textbf{100} & 0.03 & \textbf{20}\\
    \midrule
    Dom. 1, 4x4, 10ing & S (baseline) & 50 & 0.01 & 0.068\\
    \midrule
    Dom. 1, 4x4, 10ing & D (ours) & \textbf{100} & 0.04 & \textbf{16.7}\\
    \midrule[1.5pt]
    Dom. 1, 5x5, 6ing & S (baseline) & 13 & 0.01 & 0.039\\
    \midrule
    Dom. 1, 5x5, 6ing & D (ours) & \textbf{100} & 0.06 & \textbf{2.27}\\
    \midrule
    Dom. 1, 5x5, 10ing & S (baseline) & 5 & 0.01 & 0.031\\
    \midrule
    Dom. 1, 5x5, 10ing & D (ours) & \textbf{99} & 0.08 & \textbf{2.3}\\
    \midrule[1.5pt]
    Dom. 1, 6x6, 6ing & S (baseline) & 5 & 0.01 & 0.019\\
    \midrule
    Dom. 1, 6x6, 6ing & D (ours) & \textbf{100} & 0.13 & \textbf{1.15}\\
    \midrule
    Dom. 1, 6x6, 10ing & S (baseline) & 5 & 0.01 & 0.028\\
    \midrule
    Dom. 1, 6x6, 10ing & D (ours) & \textbf{97} & 0.23 & \textbf{0.338}\\
    \midrule[2pt]
    Dom. 2, 8ing & S (baseline) & 55 & 0.07 & 0.495\\
    \midrule
    Dom. 2, 8ing & D (ours) & \textbf{100} & 0.11 & \textbf{5.56}\\
    \midrule
    Dom. 2, 10ing & S (baseline) & 48 & 0.07 & 0.287\\
    \midrule
    Dom. 2, 10ing & D (ours) & \textbf{100} & 0.14 & \textbf{4.76}\\
    \bottomrule[1.5pt]
  \end{tabular}}}
\caption{Some of our experimental results. Each row reports averages
  over 100 independent episodes. Percentage of tasks solved within
  60-second timeout, belief update time (seconds), and average number
  of queries answered per second (across solved tasks) are shown. S:
  Statically factored belief (baseline), D: Dynamically factored
  belief (our method). \emph{Setting} column gives domain (``Dom. 1''
  is gridworld, ``Dom. 2'' is continuous), grid size (if applicable),
  and number of ingredients. Our approach solves more tasks than a
  static factoring does, and inference is an order of magnitude
  faster.}
\label{table:results}
\end{table}

\begin{figure}[t]
  \vspace{0.6em}
  \centering
  \noindent
  \includegraphics[width=0.58\columnwidth]{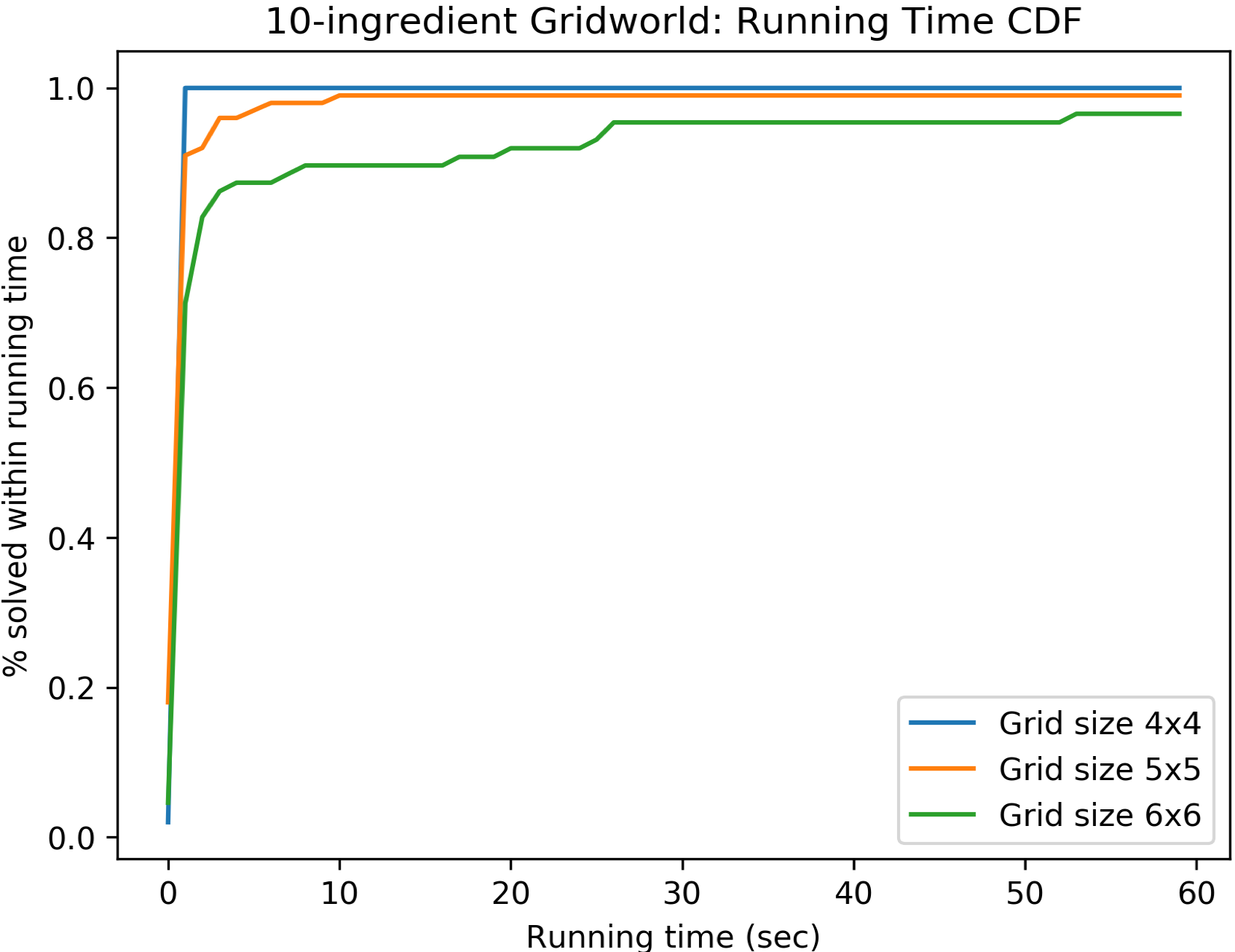}
  \includegraphics[width=0.4\columnwidth,height=3.7cm]{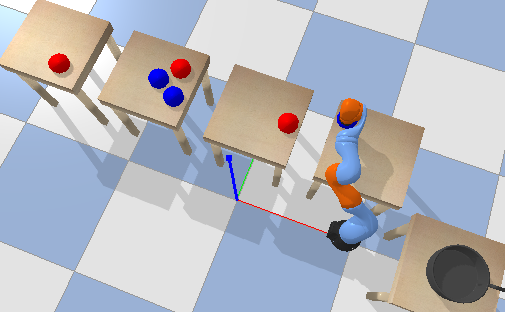}
  \caption{\emph{Left:} Cumulative distribution functions showing the
    percentage of the 100 episodes we ran with our method that got
    solved within different running times, for the 10-ingredient
    gridworld. Although our timeout was set to 60 seconds, most tasks
    were solved within 10 seconds, whereas the baseline (not shown)
    timed out frequently (see \tabref{table:results}). \emph{Right:} A visualization of the
    continuous 3D cooking domain. The robot is a light-blue-and-orange
    arm. Vegetables (red) and seasonings (blue) are placed across four
    tables.}
  \label{fig:results}
\end{figure}

\begin{figure}[t]
  \centering
    \noindent
    \includegraphics[width=0.49\columnwidth]{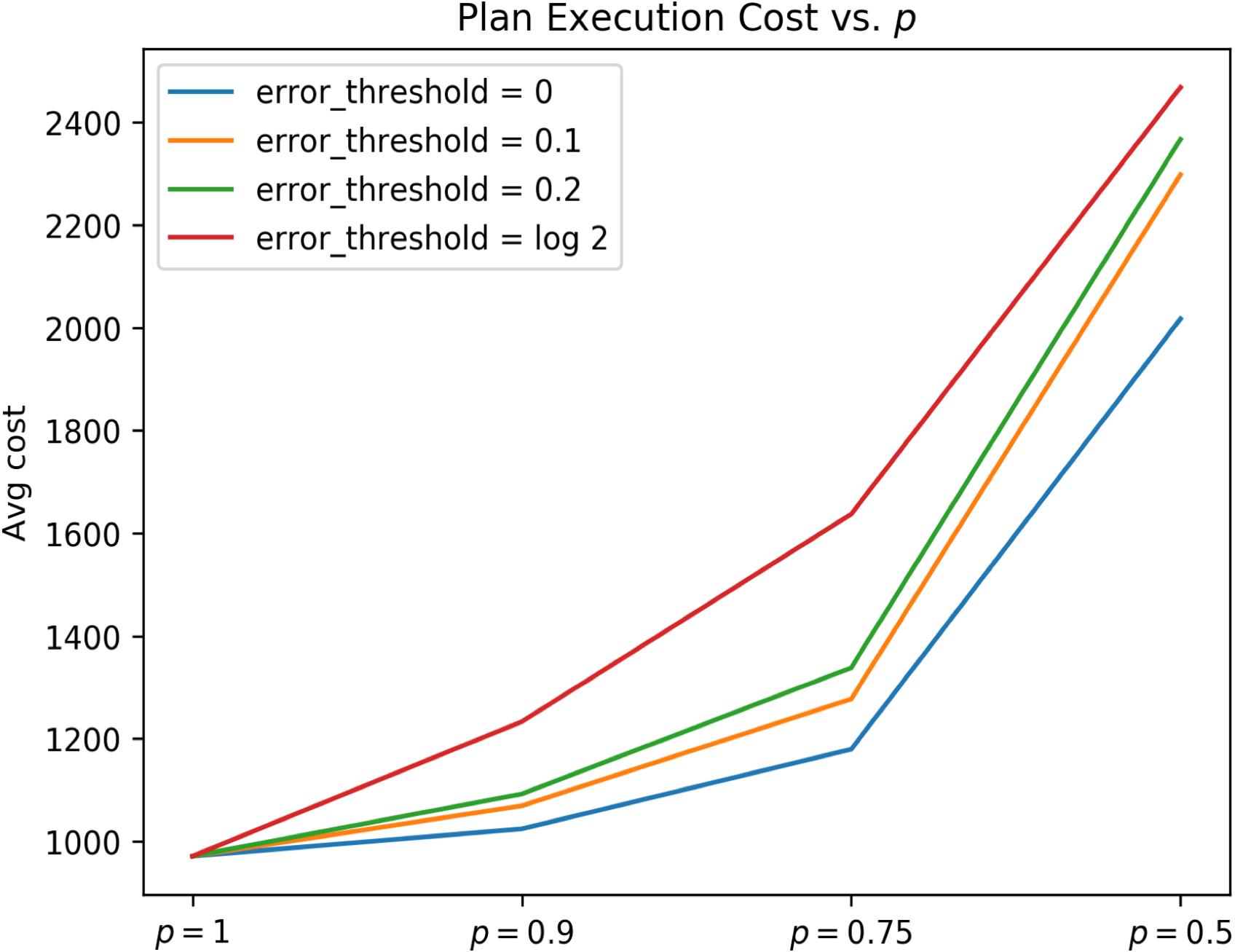}
    \includegraphics[width=0.49\columnwidth]{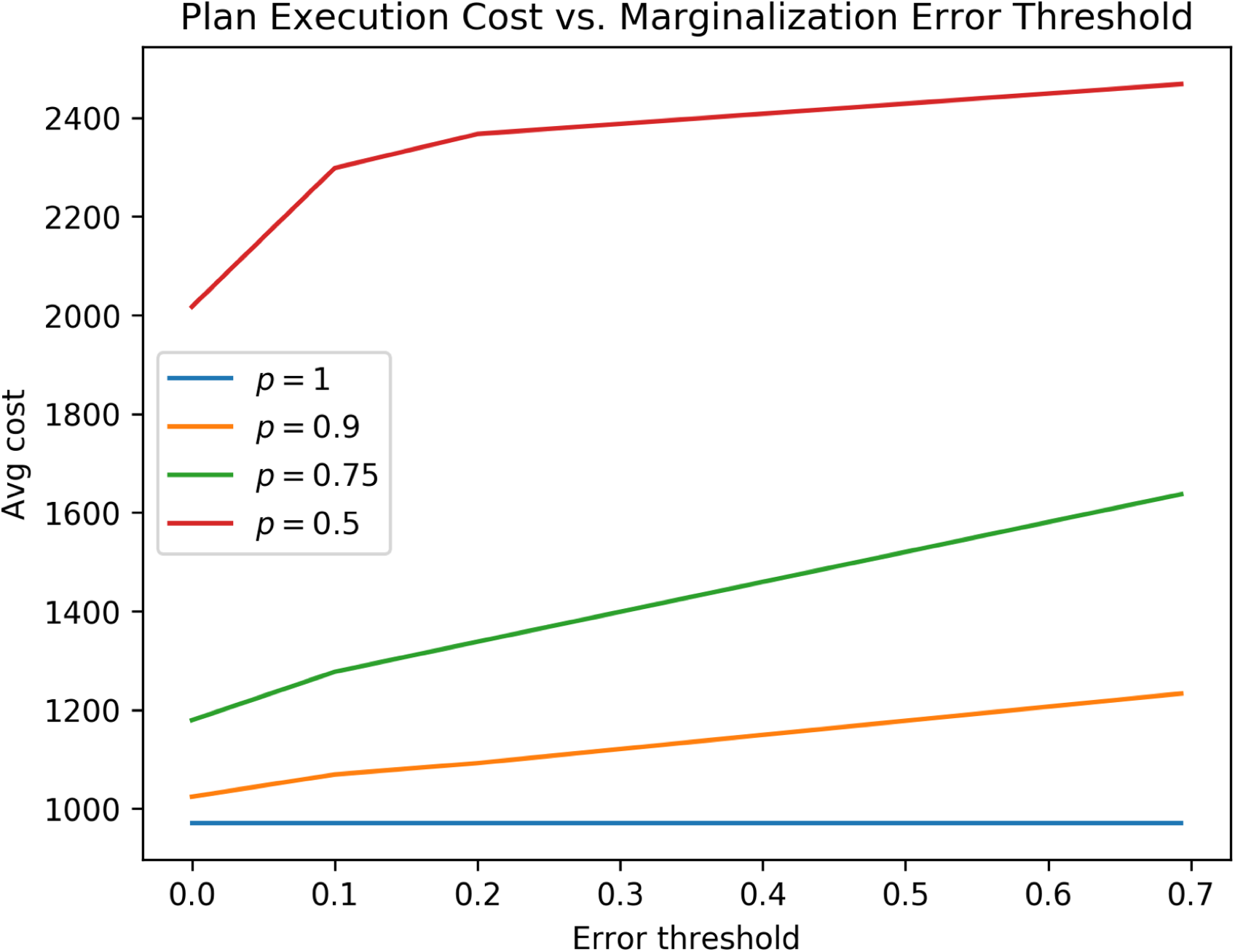}

    \vspace{0.7em}

    \includegraphics[width=0.49\columnwidth]{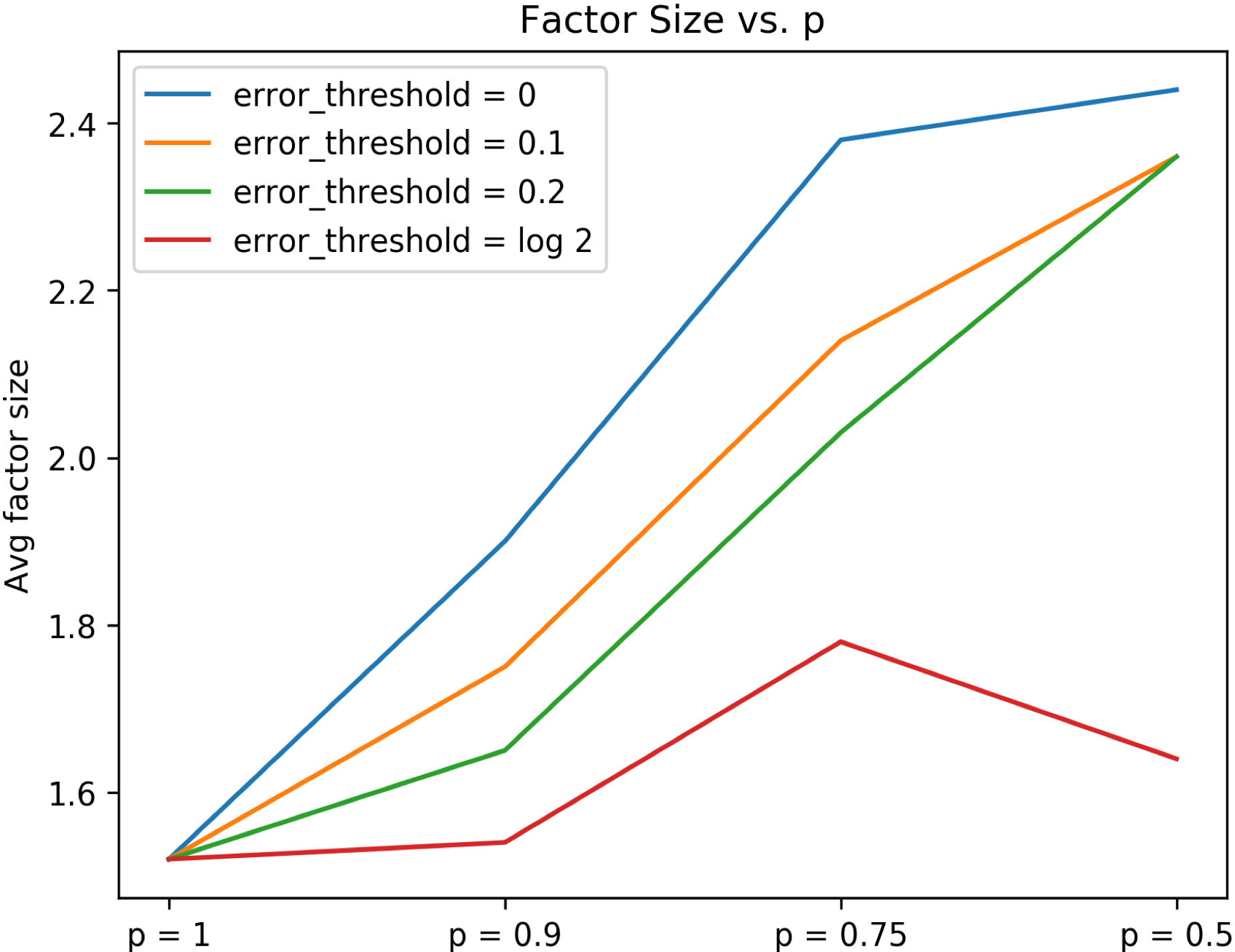}
    \includegraphics[width=0.49\columnwidth]{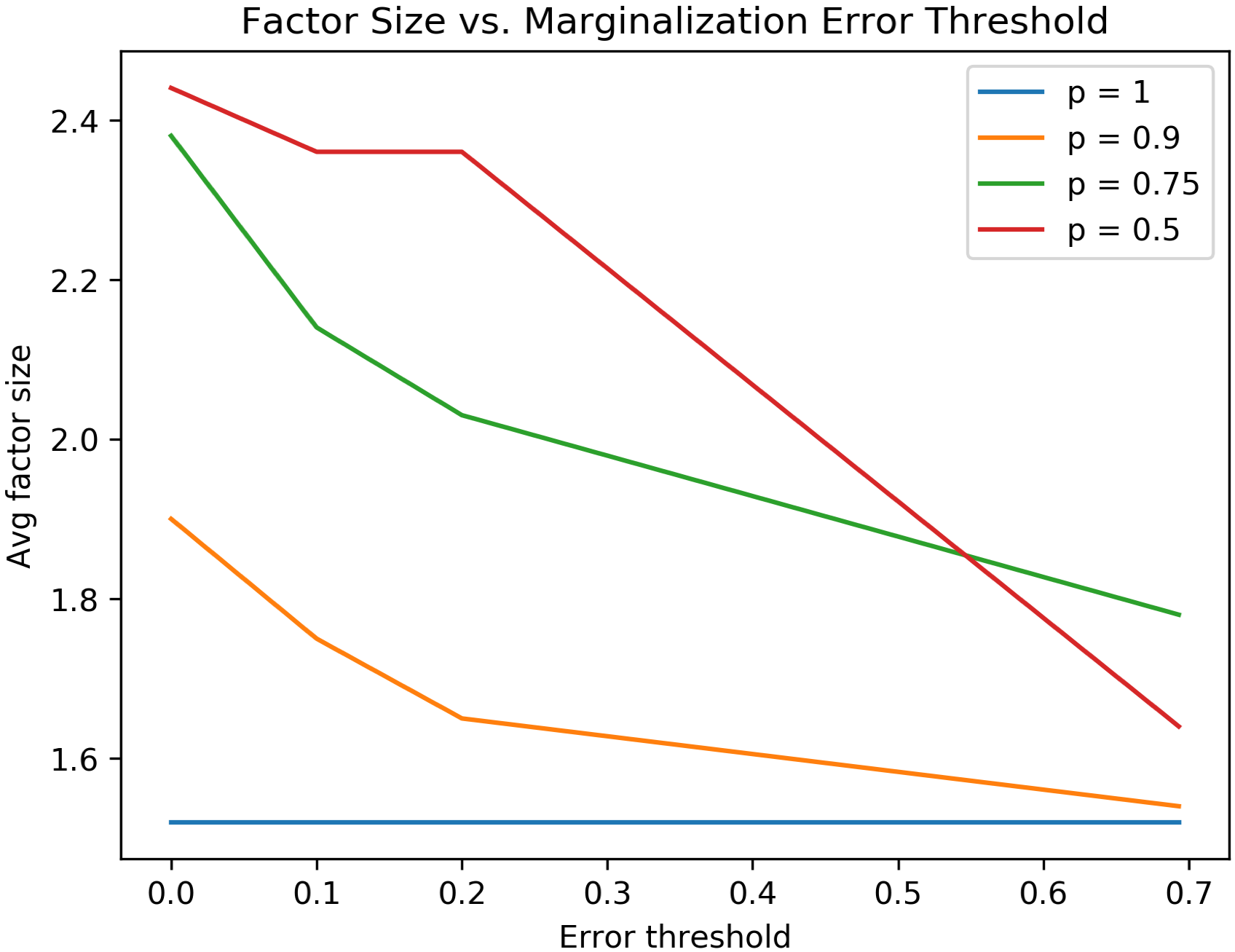}
    \caption{Results for varying $p$ and $\epsilon$
      (see \algref{alg:subroutines}) in our experiments. As $p$ decreases,
      observations get noisier, so execution costs and factor sizes
      increase. As $\epsilon$ increases, more marginalization occurs
      (with reconstruction error) and inference accuracy decreases, so
      factors are smaller but execution is costlier.}
  \label{fig:noisyresults}
\end{figure}

\textbf{Results and Discussion} \ \tabref{table:results} and
\figref{fig:results} show results when all $p$ are 1 and $\epsilon$ is
0 (see \algref{alg:subroutines}), while \figref{fig:noisyresults}
shows results with noisy observations where $p$ and $\epsilon$
vary. Overall, our approach solves significantly more tasks than a
static factoring does, and also does inference an order of magnitude
faster. However, our method could perform badly when belief
updates are very expensive, which could happen if we try to eagerly
incorporate fluents that link several state variables with large
domains. Typically in practice, though, such fluents would be placed
into ComplexFluents. As $p$ decreases, observations get
noisier, so execution costs and factor sizes increase. As $\epsilon$
increases, more marginalization occurs and inference accuracy is
lower, so factors are smaller but execution is costlier.

\section{Conclusion and Future Work}
We have considered the problem of belief state representation in an
open-domain planning problem where a human can give relational
information to the robot. We showed that a dynamically factored belief
is a good representational choice for efficient inference and planning
in this setting.

One future direction to explore is to approximately fold information
into the belief representation rather than compute a joint on every
update. We should seek a mechanism that allows the designer to trade
off between compactness of the belief and accuracy of
inference. Another direction to explore is a non-uniform observation
model: a robot given information $I$ can learn something not only from
$I$ but also from the fact that it was told $I$ as opposed to anything
else.



\section*{Acknowledgments}
We gratefully acknowledge support from NSF grants 1420316, 1523767,
and 1723381; from AFOSR grant FA9550-17-1-0165; from Honda Research;
and from Draper Laboratory. Rohan is supported by an NSF Graduate
Research Fellowship. Any opinions, findings, and conclusions expressed
in this material are those of the authors and do not necessarily
reflect the views of our sponsors.


\bibliography{references}

\end{document}